\documentclass[10pt,twocolumn,letterpaper]{article}

\usepackage[pagenumbers]{cvpr} 

\usepackage{url}

\usepackage{subcaption}
\usepackage{caption}
\usepackage{newfloat}
\usepackage{listings}
\usepackage{xspace}

\usepackage{graphbox}
\usepackage{graphicx}
\usepackage{multirow}
\usepackage{wrapfig}
\usepackage{tikz}
\usepackage{float}

\usepackage{adjustbox}
\usepackage{booktabs}
\usepackage{stackengine}
\usepackage{tablefootnote}
\usepackage{threeparttable}

\usepackage{amsmath}
\usepackage{amssymb}
\usepackage{bbm}
\usepackage{dsfont}
\usepackage{mathtools}

\usepackage{color}
\usepackage{colortbl}
\usepackage[normalem]{ulem}
\usepackage{xcolor}

\newcommand{\hfilll}{\hspace{0pt plus 1filll}}

\makeatletter

\def\OurTitle{Long-VITA: Scaling Large Multi-modal Models to 1 Million Tokens \\ with Leading Short-Context Accuracy}

\def\OurMethod{Long-VITA\xspace}

\def\OurMethodBF{\textbf{Long-VITA}\xspace}

\makeatother

\definecolor{babyblue}{rgb}{0.54, 0.81, 0.94}
\definecolor{babyblueeyes}{rgb}{0.63, 0.79, 0.95}
\makeatother

\newcommand{\checkmarknew}{\raisebox{0.6ex}{\scalebox{0.7}{$\sqrt{}$}}}
\newcommand{\crossmarknew}{\scalebox{0.85}[1]{\color{red}$\times$}}

\usepackage[normalem]{ulem}

\makeatletter

\definecolor{cvprblue}{rgb}{0.21,0.49,0.74}
\usepackage[pagebackref,breaklinks,colorlinks,allcolors=cvprblue]{hyperref}

\def\paperID{*****} 
\def\confName{CVPR}
\def\confYear{2025}

\title{\OurTitle}

\author{
    Yunhang Shen$^{1}$, Chaoyou Fu$^{2}$\thanks{Corresponding Author.}, Shaoqi Dong$^{2}$, Xiong Wang$^{1}$, Yi-Fan Zhang$^{5}$ \\
    Peixian Chen$^{1}$, Mengdan Zhang$^{1}$, Haoyu Cao$^{1}$, Ke Li$^{1}$, Shaohui Lin$^{3}$, Xiawu Zheng$^{4}$ \\
    Yan Zhang$^{4}$, Yiyi Zhou$^{4}$, Ran He$^{5}$, Caifeng Shan$^{2}$, Rongrong Ji$^{4}$, Xing Sun$^{1}$
     \\ \\
    $^{1}$Tencent Youtu Lab, $^{2}$Nanjing University \\ $^{3}$East China Normal University, $^{4}$Xiamen University, $^{5}$CASIA
    \\ \\
    \url{https://github.com/VITA-MLLM/Long-VITA}
}

\begin{document}
\maketitle
\begin{abstract}

%
%
We introduce \OurMethod, a simple yet effective large multi-modal model for long-context visual-language understanding tasks.
It is adept at concurrently processing and analyzing modalities of image, video, and text over \textbf{4K} frames or \textbf{1M} tokens while delivering advanced performances on short-context multi-modal tasks.
We propose an effective multi-modal training schema that starts with large language models and proceeds through vision-language alignment, general knowledge learning, and two sequential stages of long-sequence fine-tuning.
We further implement context-parallelism distributed inference and logit-masked language modeling head to scale \OurMethod to infinitely long inputs of images and texts during model inference.
Regarding training data, \OurMethod is built on a mix of $17$M samples from \textbf{public datasets only} and demonstrates state-of-the-art performance on various multi-modal benchmarks, compared against recent cutting-edge models with internal data.
%
%
\OurMethod is fully open-source and reproducible.
By leveraging our inference designs, \OurMethod models achieve a remarkable \textbf{2$\times$ prefill speedup} and \textbf{4$\times$ context length extension} in a single node with $8$ GPUs.
%
We hope \OurMethod can serve as a competitive baseline and offer valuable insights for the open source community to advance long-context multimodal understanding.

\end{abstract}

\section{Introduction}
\label{sec:intro}

In recent years, proprietary Large Multi-Modal Models (LMMs) have been undergoing rapid iteration and evolution~\cite{GPT-4,Gemini2.0,Claude3.5}, progressively extending large language models (LLMs) with multi-sensory skills, such as visual understanding.
Beyond closed-source models, open-source models, including LLaVA series~\cite{LLaVA-1.5,LLaVA-OneVision}, Qwen-VL series~\cite{Qwen-VL,Qwen2-VL}, and VITA series~\cite{VITA,VITA-1.5}, are also making significant strides, trying to close the gap with their closed-source counterparts.
However, most of the above open-source works on visual understanding tasks typically focus on static images and short video inputs.
Proprietary models show superior support for long-content inputs, while open-source models lag significantly behind.
For example,  Gemini 1.5 pro~\cite{Gemini1.5} runs up to 1 million tokens in production and processes 1 hour of video information in one go.
Therefore, more information is needed on developing high-performing, long-context vision-language models for public use.
Recently, many works~\cite{LongVILA,LongLLaVA,LongVU,LongVA} have been proposed to address the challenges of training and inference of long-context information.
However, they~\cite{LongVILA,LongLLaVA} aim mainly to improve a comprehensive understanding of long video, neglecting the static image and short video input scenarios.
On the other hand, some works~\cite{SF-LLaVA,LongVU} rely on compressing visual tokens, which often comes at the expense of performance degradation.
To further push the limits of open-source model capabilities, we extend the context length to $1$ million tokens and introduce \OurMethod, a strong open-source long-context visual language model.
To this end, we employ a phased training approach that positions language as the pivot.
Specifically, \OurMethod’s ability to handle extended contexts is systematically augmented through a four-stage process.
Beyond the conventional stages of vision-language alignment and supervised fine-tuning, our approach incorporates specialized stages for long-context supervised fine-tuning of $128$K and $1$M.
To achieve a good performance trade-off between long and short sequences, \OurMethod takes advantage of the existing abundance of open-source image-text and video-text data.
We also introduce a multi-image summarization dataset, Comic-9K, comprising $9$k comic books and the corresponding detailed synopsis.
This dataset has a total of $200$k images with an average of $20$ high-resolution photos for each sample, and the synopsis is all manually written and collected from the web.

Our contributions can be summarized as follows.
\begin{itemize}
\item
We fully release \OurMethod to the open-source community, providing a powerful tool for developing and applying long-context multi-modal AI systems and encouraging further research in this domain.

\item
We further implement context-parallelism distributed inference and logits-masked language modeling head to scale up the infinite number of image and text tokens for model deployment.
%


\item 
We evaluate \OurMethod on a wide range of benchmarks.
Although \OurMethod is trained only on open-source data, comprehensive evaluations reveal that \OurMethod has emerged as a strong vision-language model among previous models of similar scale, especially in evaluating hallucinations and video understanding.

\end{itemize}

\section{Related Work}
\label{sec:related_work}

\subsection{Large Vision Language Models}
Recent advancements have seen the creation of large vision language models (LVLMs), which usually enhance large language models with the capability to process and interpret visual information.
Flamingo~\cite{Flamingo} performs various multi-modal tasks, \eg, image captioning, visual dialogue, classification, or visual question answering, from only a few input/output examples.
BLIP-2~\cite{BLIP-2} leverages frozen pre-trained image encoders and large language models to bootstrap vision-language pre-training.
LLaVA~\cite{LLaVA} uses language models to generate multi-modal instruction-following data and connects a vision encoder and large language models for general-purpose visual and language understanding.
Qwen-VL~\cite{Qwen-VL} and InternVL~\cite{InternVL} series further perform various vision-language tasks, such as image captioning, question answering, text-oriented question answering, and visual grounding.
These works showcase that LVLMs have achieved significant breakthroughs.
Furthermore, significant progress in multi-modal model evaluation~\cite{MMBench,MME,MMStar} has also contributed to the rapid improvement of large vision-language models.
In this work, we introduce \OurMethod, a series of multi-modal and long-content models trained exclusively with fully open-source datasets for pre-training and supervised fine-tuning, demonstrating promising results on extensive benchmarks.

\subsection{Long-Context Multi-Modal Model}
LLMs are typically pre-trained with a pre-defined context length.
Training LLMs with long context from scratch is prohibitively expensive for most researchers.
Recently, several works, \eg, Position Interpolation~\cite{PositionInterpolation}, YaRN~\cite{YaRN}, LongRoPE~\cite{LongRoPE} and LongLoRA~\cite{LongLoRA} have tried to extend the context length of LLMs by fine-tuning.
Many methods have also been proposed in large multi-modal models to handle long-context visual inputs.
LongVILA~\cite{LongVILA} executes a continuation of pre-training on the LLM to enhance its context length to $256$K, followed by long video training.
LongLLaVA~\cite{LongLLaVA} integrates a hybrid of Mamba and Transformer blocks for long-context multi-modal understanding.
LongVU~\cite{LongVU} proposes a spatio-temporal adaptive compression scheme to reduce long video tokens by leveraging cross-modal query and inter-frame similarities.
LongVA~\cite{LongVA} extends the language model on text data and then aligns the extended model with visual inputs, which perceives more than $200$K visual tokens.
Kangaroo~\cite{Kangaroo} develops a curriculum training strategy that progressively equips the LLM with the capacity to comprehend long videos.
However, the above methods mainly focus on video understanding and visual information retrieval, neglecting the trade-off between image and long-video understanding tasks.
In this paper, we propose \OurMethod, a powerful LMM that obtains a long-context capacity of $1$ million tokens and simultaneously achieves superior performance on both image and video understanding tasks.

\subsection{Long-Context Visual Instruction Data}
LLaVA-Video~\cite{LLaVA-Video} synthesizes long video-language instruction data, covering various tasks such as captioning, open-ended, and multi-choice QA.
LongVILA~\cite{LongVILA} constructs instruction-following datasets from long videos, which encompass summarization and other queries relevant to a comprehensive understanding of the content of long videos.
Video-MME~\cite{Video-MME} incorporates a diverse range of video types, varying temporal durations, ranging from $11$ seconds to $1$ hour.
LVBench~\cite{LVBench} evaluates long-context LMMs that feature video-language interleaved inputs up to an hour long.
LongVideoBench~\cite{LongVideoBench} introduces referring reasoning questions and presents significant challenges for both proprietary and open-source LMMs in their long-context multi-modal capabilities.
However, the above works only focus on long-context understanding in videos, which usually contain redundant visual frames and other modal information, such as subtitles and audio.
In this paper, we explore the comic-based long-context instruction learning and collect a high-quality real dataset for comic book summarization.

\section{\OurMethod}

\subsection{Architecture}
\label{sec:Architecture}

\begin{table*}[!htbp]
 \caption{
 Summary of datasets used in \OurMethod for different stages.
 `Comic-9K' and `MovieNet-Summary' are publicly available at:
 \href{https://huggingface.co/datasets/VITA-MLLM/Comic-9K}{https://huggingface.co/datasets/VITA-MLLM/Comic-9K} and \href{https://huggingface.co/datasets/VITA-MLLM/MovieNet-Summary}{https://huggingface.co/datasets/VITA-MLLM/MovieNet-Summary}, respectively.
 }
 \begin{center}
 \begin{adjustbox}{max width=0.99\textwidth}
 \begin{tabular}{c|lc|cccc}
 \toprule

 \multirow{2}{*}{Type} 
 & \multirow{2}{*}{Name}
 & \multirow{2}{*}{Total Number} & \multicolumn{4}{c}{Sampling Ratio / Max Number} \\
 
 & & & Stage 1 & Stage 2 & Stage 3 & Stage 4\\

 \midrule

 \multirow{9}{*}{Image-Text}
    
 & LLaVA-ReCap~\hfilll~\cite{li2024llavanext-ablations} & 3.5M & 1.0 & 1.0 & 0.1 & 0.1 \\

 & ALLaVA-4V~\hfilll~\cite{ALLaVA-4V} & 1.4M & 1.0 & 1.0 & 0.1 & 0.1 \\

 & LVIS-Instruct4V~\hfilll~\cite{LVIS-Instruct4V} & 222K & 0.0 & 1.0 & 0.1 & 0.1 \\

 & ShareGPT4V~\hfilll~\cite{ShareGPT4V} & 1.3M & 1.0 & 1.0 & 0.1 & 0.1 \\

 & the-cauldron~\hfilll~\cite{Idefics2} & 1.1M & 0.0 & 1.0 & 0.1 & 0.1 \\

 & Docmatix~\hfilll~\cite{Idefics3} & 1.2M & 1.0 & 1.0 & 0.1 & 0.1 \\

 & LLaVA-OneVision-Mid~\hfilll~\cite{LLaVA-OneVision} & 444K & 1.0 & 1.0 & 0.1 & 0.1 \\

 & LLaVA-OneVision~\hfilll~\cite{LLaVA-OneVision} & 3.6M & 0.0 & 1.0 & 0.1 & 0.1 \\

 & M4-Instruct~\hfilll~\cite{li2024llavanext-interleave} & 860K & 0.0 & 1.0 & 0.1 & 0.1 \\

 & Comic-9K & 9K & 0.0 & 0.0 & 1.0 & 1.0 \\

 \midrule

 \multirow{5}{*}{Video-Text}
 
 & VideoGPT-plus~\hfilll~\cite{VideoGPT-plus} & 575K & 0.0 & 1.0 & 0.1 & 0.1 \\

 & ShareGemini-cap~\hfilll~\cite{ShareGemini} & 323K & 0.0 & 1.0 & 0.1 & 0.1 \\

 & LLaVA-Video-178K~\hfilll~\cite{LLaVA-Video} & 1.6M & 0.0 & 0.0 & 1.0 & 1.0 \\
 
 & MovieNet-Summary & 1K & 0.0 & 0.0 & 0.0 & 1.0 \\
 
 \midrule

 \multirow{9}{*}{Short Text}
 
 & OpenHermes-2.5~\hfilll~\cite{OpenHermes-2.5} & 1.0M & 0.0 & 1.0 & 0.1 & 0.1 \\

 & LIMA~\hfilll~\cite{LIMA} & 1K & 0.0 & 1.0 & 0.1 & 0.1 \\

 & databricks-dolly-15k~\hfilll~\cite{databricks-dolly-15k} & 15K & 0.0 & 1.0 & 0.1 & 0.1 \\

 & MetaMathQA~\hfilll~\cite{MetaMathQA} & 395K & 0.0 & 1.0 & 0.1 & 0.1 \\

 & MathInstruct~\hfilll~\cite{MathInstruct} & 262K & 0.0 & 1.0 & 0.1 & 0.1 \\

 & Orca-Math~\hfilll~\cite{Orca-Math} & 200K & 0.0 & 1.0 & 0.1 & 0.1 \\

 & atlas-math-sets~\hfilll~\cite{atlas-math-sets} & 17.8M & 0.0 & 1.0 & 0.1 & 0.1 \\

 & goat~\hfilll~\cite{goat} & 1.7M & 0.0 & 1.0 & 0.1 & 0.1 \\

 & camel-ai-math~\hfilll~\cite{CAMEL} & 50K & 0.0 & 1.0 & 0.1 & 0.1 \\

 \midrule

 \multirow{8}{*}{Long Text}
 
 & Long-Instruction~\hfilll~\cite{Long-Instruction-with-Paraphrasing} & 16K & 0.0 & 0.0 & 1.0 & 1.0 \\

 & LongForm~\hfilll~\cite{LongForm} & 23K & 0.0 & 0.0 & 1.0 & 1.0 \\

 & LongAlign-10k~\hfilll~\cite{LongAlign} & 10K & 0.0 & 0.0 & 1.0 & 1.0 \\

 & LongCite-45k~\hfilll~\cite{LongCite} & 45K & 0.0 & 0.0 & 1.0 & 1.0 \\

 & LongWriter-6k~\hfilll~\cite{LongWriter} & 6K & 0.0 & 0.0 & 1.0 & 1.0 \\

 & LongQLoRA~\hfilll~\cite{LongQLoRA} & 39K & 0.0 & 0.0 & 1.0 & 1.0 \\

 & LongAlpaca~\hfilll~\cite{LongLoRA} & 12K & 0.0 & 0.0 & 1.0 & 1.0 \\

 & Long-Data-Collections~\hfilll~\cite{Long-Data-Collections} & 98K & 0.0 & 0.0 & 1.0 & 1.0 \\

 \bottomrule
 \end{tabular}
 \end{adjustbox}
 \end{center}
 \label{table_data}
\end{table*}

\begin{table*}[!htbp]
 \caption{
 Detailed configuration for each training stage of the \OurMethod model.
 }
 \begin{center}
 \begin{adjustbox}{max width=0.99\textwidth}
 \begin{tabular}{c|c|c|c|c|c}
 \toprule

 \multicolumn{2}{c|}{} & Stage 1 & Stage 2 & Stage 3 & Stage 4\\

\midrule
\multicolumn{2}{c|}{Sequence Length} & 32K & 16K & 128K & 1M \\
\midrule
\multicolumn{2}{c|}{Batch Size} & 528 & 528 & 64 & 8 \\

\midrule
\multicolumn{2}{c|}{Training Iterations} & 1,000 & 5,000 & 1,000 & 500 \\
\midrule
\multicolumn{2}{c|}{Training Tokens} & 16B & 40B & 8B & 4B \\
\midrule
\multicolumn{2}{c|}{Sequence Packing} & \checkmarknew & \checkmarknew & \checkmarknew & \checkmarknew \\

\midrule
\multirow{4}{*}{Parallelism}
& \multicolumn{1}{c|}{Tensor} & 8 & 8 & 8 & 8 \\
& \multicolumn{1}{c|}{Pipeline} & 1 & 1 & 1 & 1 \\
& \multicolumn{1}{c|}{Context} & 1 & 1 & 2 & 8 \\
& \multicolumn{1}{c|}{Data} & 8 & 8 & 4 & 1 \\

\midrule
\multirow{2}{*}{Image} & Resolution & \multicolumn{4}{c}{$448 + 448 \times \{1\times2, 2\times1, \dots, 3\times4, 4\times3\} $} \\
\cmidrule{2-6}
&Max Number & 128 & 64 & 512 & 4,096 \\
\midrule
\multirow{3}{*}{Video} & Resolution & \multicolumn{4}{c}{$448$} \\
\cmidrule{2-6}
&FPS & \multicolumn{4}{c}{$1$} \\
\cmidrule{2-6}
&Max Frames & 128 & 64 & 512 & 4,096 \\
\midrule

\multirow{3}{*}{Learing Rate} & Vision & $0.0$ & $1.0 \times 10^{-6}$ & $1.0 \times 10^{-6}$ & $1.0 \times 10^{-6}$ \\
\cmidrule{2-6}
& Projector & $1.0 \times 10^{-3}$ & $1.0 \times 10^{-5}$ & $1.0 \times 10^{-5}$ & $1.0 \times 10^{-5}$ \\
\cmidrule{2-6}
& LLM & $0.0$ & $1.0 \times 10^{-5}$ & $1.0 \times 10^{-5}$ & $1.0 \times 10^{-5}$ \\
\midrule
\multirow{1}{*}{Learing Rate Decay} & Vision & $0.0$ & \multicolumn{3}{c}{$0.9$} \\
\midrule
\multicolumn{2}{c|}{\multirow{1}{*}{Learing Rate scheduler}} & \multicolumn{4}{c}{Cosine} \\
\midrule
\multicolumn{2}{c|}{Weight Decay} & \multicolumn{4}{c}{$0.0$} \\
\midrule
\multicolumn{2}{c|}{Gradient Clip} & \multicolumn{4}{c}{$1.0$} \\
\midrule
\multicolumn{2}{c|}{Rotary Base} & \multicolumn{4}{c}{$1,000,000$} \\
\midrule
\multicolumn{2}{c|}{Adam Beta1} & \multicolumn{4}{c}{$0.9$} \\
\midrule
\multicolumn{2}{c|}{Adam Beta2} & \multicolumn{4}{c}{$0.999$} \\

 \bottomrule
 \end{tabular}
 \end{adjustbox}
 \end{center}
 \label{table_train}
\end{table*}

This technical report presents \OurMethod, a new series of open-source vision-language models that explore long-context vision understanding without token compression and sparse local attention.
Our multi-modal architecture is constructed around three core components: the Vision Encoder, the Projector, and the LLM.
%

    \textbf{Large Language Model.}
    We choose Qwen2.5-14B-Instruct~\cite{Qwen2.5} as our LLM.
    
    \textbf{Vision Encoder.}
    We consider the InternViT-300M~\cite{InternVL} as the visual encoder.
    We introduce a dynamic tiling vision encoding strategy~\cite{InternVL2} that efficiently processes high-resolution images of varying aspect ratios.

    \textbf{Vision-Language Projector.}
    We employ a $2$-layer MLP to project image features into the word embedding space.
    We also apply a simple pixel shuffle~\cite{InternVL2} to visual tokens and reduce the number of visual tokens to one-quarter.
    %

\subsection{Data Construction}
\label{sec:Data}

\OurMethod is trained on open-source datasets only.
As shown in Tab.~\ref{table_data}, the training dataset encompasses a diverse range of sources.

\textbf{Image-Text Data.}
The datasets employed can be categorized into three groups:
\begin{itemize}[leftmargin=2.0em]
    \item
    \textbf{Image Captioning.}
    The visual caption dataset consists of LLaVA-ReCap~\cite{li2024llavanext-ablations}, ALLaVA-4V~\cite{ALLaVA-4V}, ShareGPT4V~\cite{ShareGPT4V} and LLaVA-OneVision-Mid~\cite{LLaVA-OneVision}.

    \item
    \textbf{Visual Question Answering.}
    We combine general VQA from LVIS-Instruct4V~\cite{LVIS-Instruct4V}, the-cauldron~\cite{Idefics2}, Docmatix~\cite{Idefics3}, LLaVA-OneVision~\cite{LLaVA-OneVision}.

    \item
    \textbf{Interleaved Image-Text.}
    To empower all-round multi-image capabilities, we employ the M4-Instruct~\cite{li2024llavanext-interleave}.
    To further enhance multi-image understanding with more than $10$ images, we collect the public comic book with the corresponding detailed synopsis from the web and build the Comic-9k datasets.
    Specifically, Comic-9k contains $ 200,000$ images, spanning $ 9,000$ comic books, along with a manually labeled synopsis.
\end{itemize}

\begin{table*}[!htbp]
 \caption{
 Maximal supported sequence length for inference and training.
 }
 \begin{center}
  \begin{adjustbox}{max width=0.99\textwidth}
   \begin{tabular}{lcc|ccc|c}
    \toprule

     \multirow{3}{*}{Name} & \multirow{3}{*}{\# Param.} & \multirow{3}{*}{Device Type} & \multicolumn{4}{c}{Devices Number} \\
     \cmidrule{4-7}
     &&& \multicolumn{3}{c|}{Inference} & Training \\
     \cmidrule{4-7}
     &&& 8 & 16 & 32 & 64 \\

    \midrule
    
    LongVILA~\cite{LongVILA}            & 7B & 80G GPU & 276K & -- & -- & 666K \\
    
    \midrule
    \multirow{2}{*}{\OurMethodBF}       & \multirow{2}{*}{14B} & 64G NPU & -- & 1,024K & -- & 1,024K \\
    
                                        & & 96G GPU  & 400K & 800K & 1,600K & 1,024K \\

    \bottomrule
   \end{tabular}
  \end{adjustbox}
 \end{center}
 \label{table_max_length}
\end{table*}

\begin{table*}[!htbp]
	\caption{
		Comparison of different language modeling head.
	}
	\begin{center}
		\begin{adjustbox}{max width=0.99\textwidth}
			\begin{tabular}{c|r|rl|l}
				\toprule

				\# GPUs & Method & \# Frames & \# Tokens & Prefill Time (s) \\
				
				\midrule
				\multirow{5}{*}{8}
				
				&\multirow{2}{*}{Original LM Head}
				& 390 & 100K & 19.2 \\
				
				&& 400 & 103K & OOM \\

				\cmidrule{2-5}
				
				& \multirow{3}{*}{Logits-Masked LM Head}
				& 390 & 100K & 10.1 \color{red}{ ( $\downarrow$ 47.3\% ) } \\
				&& 1,620 & 417K \color{red}{ ( $\uparrow$ 417\% ) } & 102 \\
				&& 1,630 & 420K & OOM \\
				
				\midrule
				
				\multirow{4}{*}{32}
				
				& \multirow{2}{*}{Chunked LM Head}
				& 4,096 & 1,056K & 177 \\
				&& 6,400 & 1,651K & 377 \\
				
				\cmidrule{2-5}
				
				& \multirow{2}{*}{Logits-Masked LM Head}
				& 4,096 & 1,056K & 157 \color{red}{ ( $\downarrow$ 11.3\% ) } \\
				&& 6,400 & 1,651K & 335 \color{red}{ ( $\downarrow$ 11.1\% ) } \\

				\bottomrule
			\end{tabular}
		\end{adjustbox}
	\end{center}
	\label{table_lm_head}
\end{table*}

\textbf{Video-Text Data.}
We construct our video understanding data using VideoGPT-plus~\cite{VideoGPT-plus}, ShareGemini~\cite{ShareGemini}, and LLaVA-Video-178K~\cite{LLaVA-Video}.
To improve the long-context capability of movie-level video understanding, we build a MovieNet-Summary dataset, which consists of paired movies and synopses from MovieNet~\cite{MovieNet}.
\textbf{Short Text Data.}
Following~\cite{Idefics3}, the pure text data is collected from OpenHermes-2.5~\cite{OpenHermes-2.5}, LIMA~\cite{LIMA}, databricks-dolly-15k~\cite{databricks-dolly-15k}, MetaMathQA~\cite{MetaMathQA}, MathInstruct~\cite{MathInstruct}, Orca-Math~\cite{Orca-Math}, atlas-math-sets~\cite{atlas-math-sets}, goat~\cite{goat}, and camel-ai-math~\cite{CAMEL}.
\textbf{Long Text Data.}
To transfer the context length of the language model to the modality-aligned multi-modal models~\cite{LongVA}, we gather several long text datasets, including Long-Instruction-with-Paraphrasing~\cite{Long-Instruction-with-Paraphrasing}, LongForm~\cite{LongForm}, LongAlign-10k~\cite{LongAlign}, LongCite-45k~\cite{LongCite}, LongWriter-6k~\cite{LongWriter}, LongQLoRA~\cite{LongQLoRA}, LongAlpaca~\cite{LongLoRA}, and Long-Data-Collections~\cite{Long-Data-Collections}.
Note that Comic-9k and MovieNet-Summary were created by this work and are made publicly available.
Therefore, \OurMethod is \textbf{only} trained on open data, and we \textbf{do not} use data filtering methods.

\subsection{Training Pipelines}
\label{sec:Training}

Unlike other models, \OurMethod training is divided into four stages with varying sequence lengths.
%

    \textbf{Stage 1: Vision-Language Alignment.}
    Building upon pre-trained language models, our primary objective is to establish initial connections between visual features and language features.
    We freeze the LLM and the visual encoder, only training the visual projector.
    Therefore, we mainly use caption data for pre-training.
    We also add Docmatix~\cite{Idefics3} in this stage to improve document-based VQA.
    %

    \textbf{Stage 2: Learning of general knowledge.}
    After establishing vision-language alignment in the embedding space, we dedicate most of our computational resources to general knowledge learning in the vision-language.
    This stage leverages all the image-text data for multiple tasks, including image captioning, common VQA, OCR, and multi-model conversations.
    In this stage, we also add text-only general instructions, math problems, and arithmetic calculations.
    For video understanding, we only add VideoGPT-plus~\cite{VideoGPT-plus} and ShareGemini-cap~\cite{ShareGemini}.
    In both Stage 1 and Stage 2, we pack all training data to a fixed sequence length, which effectively trains samples with different lengths of sequences.
    Specifically, we randomly sample data items from the same source and concatenate them into one training sample with a token length of $32$K and $16$K for Stage 1 and Stage 2, respectively.
    We reset positional embeddings and attention masks for all packed samples so that each text-vision pair only attends to itself.
    This approach helps manage extensive datasets and ensure the coverage of diverse data segments.
    %

    \textbf{Stage 3: Long-Sequence Fine-Tuning.}
    In this stage, we extend the context length to $128$K.
    We reduce the sampling ratio of the data in Stage 2 to $0.1$ and incorporate additional long-context text instructions, comic book summaries, and video understanding datasets.
    %

    \textbf{Stage 4: Long-Sequence Fine-Tuning.}
    In this stage, we extend the context length to $1,024$K and add additional movie summary data.
    In both Stage 3 and Stage 4, we also pack all training data to a fixed sequence length without resetting positional embedding and attention mask.
    Therefore, we impose the model to capture the correlations between these two modalities in long-contextual information.
    %
    

%
We \textbf{do not} use the interpolation technique during training and testing; therefore, the context window of \OurMethod can be extended further when equipped with YaRN~\cite{YaRN}, LongRoPE~\cite{LongRoPE}, and NTK-based interpolation.
Note that we \textbf{do not} use any parameter-efficient methods such as LoRA~\cite{LoRA} or approximate attention~\cite{LongLoRA}.

\begin{table*}[!htbp]
 \caption{
 Comparison with the state-of-the-art models under \textbf{20B} parameters on OpenCompass Leaderboard.
 MMB: the test split of MMBench~\cite{MMBench}, MV: MathVista~\cite{MathVista}, HB: HallusionBench~\cite{HallusionBench}, OCR: OCRBench~\cite{OCRBench}.
 `AVG-6' denotes the average scores of six \textbf{objective} benchmarks, \ie, MMBench, MMStar, MMMU, HallusionBench, AI2D, and OCRBench, which do not use a judge LLM to evaluate.
 `AVG' denotes the average of scores on all eight benchmarks.
 `Internal Data' denotes whether the model is trained with in-house data, which is not publicly available. 
 Results are obtained from the leaderboard of OpenCompass.
 }
 \begin{center}
 \begin{adjustbox}{max width=0.99\textwidth}
 \begin{tabular}{lc|ccccc}
 \toprule

 Name
 & Internal Data
 & MMB & MMStar & MMMU & MV & HB \\
 
 \midrule
 \multicolumn{7}{c}{Open weight models \& Partially open models} \\

 Qwen2-VL-7B~\hfilll~\cite{Qwen2-VL} & \checkmarknew & 81.0 & 60.7 & 53.7 & 61.4 & 50.4 \\

 InternVL2-8B~\hfilll~\cite{InternVL2} & \checkmarknew & 79.5 & 61.5 & 51.2 & 58.3 & 45.0 \\
 InternVL2.5-8B~\hfilll~\cite{InternVL2.5} 	 & \checkmarknew & \textbf{83.2} & 62.8 & 56.0 & 64.4 & 50.1 \\

 LLaVA-OneVision-7B~\hfilll~\cite{LLaVA-OneVision}	& \checkmarknew & 80.9 & 61.9 & 47.9 & 62.3 & 31.6 \\

 POINTS1.5-7B~\hfilll~\cite{POINTS1.5} & \checkmarknew & 80.7 & 61.1 & 53.8 & 66.4 & 50.0 \\

 Ovis1.5-Gemma2-9B~\hfilll~\cite{Ovis} & \crossmarknew & 77.3 & 58.1 & 49.7 & 65.6 & 48.2 \\
 Ovis1.6-Gemma2-9B~\hfilll~\cite{Ovis} & \checkmarknew & 80.5 & \textbf{62.9} & 55.0 & \textbf{67.2} & 52.2 \\

 Llama-3.2-11B-Vision-Instruct~\hfilll~\cite{Llama-3.2} & \checkmarknew & 65.8 & 49.8 & 48.0 & 48.6 & 40.3 \\

 Pixtral-12B~\hfilll~\cite{Pixtral-12B}	 & \checkmarknew & 72.7 & 54.5 & 51.1 & 56.9 & 47.0 \\

 OmChat-v2.0-13B~\hfilll~\cite{OmChat} & \checkmarknew & 79.5 & 58.2 & 49.6 & 57.1 & 48.4 \\
 
 bailingMM-mini-17B~\hfilll~\cite{bailingmm} & \checkmarknew & 82.2 & 61.3 & 50.0 & 70.5 & 45.4 \\
 
 CogVLM2-19B-Chat~\hfilll~\cite{CogVLM2}		& \checkmarknew & 70.7 & 50.5 & 42.6 & 38.6 & 41.3 \\

 VITA-1.5-7B~\hfilll~\cite{VITA-1.5}          & \checkmarknew & 76.8 & 60.2 & 52.6 & 66.2 & 44.6 \\	
 
 \midrule
 \multicolumn{7}{c}{Fully open models} \\

 VILA1.5-13B~\hfilll~\cite{VILA} & \crossmarknew & 68.5 & 44.2 & 41.1 & 42.5 & 39.3 \\

 \OurMethodBF\textbf{-16K} & \crossmarknew & 79.8 & 61.3 & \textbf{57.0} & 65.3 & \textbf{64.6 }\\

 \OurMethodBF\textbf{-128K} & \crossmarknew & 79.5 & 60.5 & 56.7 & 65.5 & \textbf{64.6} \\
 
 \OurMethodBF\textbf{-1M }& \crossmarknew & 75.0 & 53.0 & 51.0 & 50.3 & 58.7 \\

\bottomrule

\end{tabular}
\end{adjustbox}
\bigskip

\begin{adjustbox}{max width=0.99\textwidth}
\begin{tabular}{lc|ccc|cc}
 \toprule

 Name
 & Internal Data
 & AI2D & OCR & MMVet & {AVG} & {AVG-6} \\
 
 \midrule
 \multicolumn{7}{c}{Open weight models \& Partially open models} \\

 Qwen2-VL-7B~\hfilll~\cite{Qwen2-VL} & \checkmarknew & 83.0 & \textbf{843} & 61.8 & 67.0 & 68.9 \\

 InternVL2-8B~\hfilll~\cite{InternVL2} & \checkmarknew & 83.6 & 794 & 54.3 & 64.1 & 66.7 \\
 InternVL2.5-8B~\hfilll~\cite{InternVL2.5} 	 & \checkmarknew & \textbf{84.5} & 822 & 62.8 & \textbf{68.2} & \textbf{69.8} \\

 LLaVA-OneVision-7B~\hfilll~\cite{LLaVA-OneVision}	& \checkmarknew & 82.4 & 622 & 51.9 & 60.1 & 61.1 \\

 POINTS1.5-7B~\hfilll~\cite{POINTS1.5} & \checkmarknew & 81.4 & 823 & 62.2 & 67.2 & 68.2 \\

 Ovis1.5-Gemma2-9B~\hfilll~\cite{Ovis} & \crossmarknew & \textbf{84.5} & 752 & 53.8 & 64.1 & 65.5 \\
 Ovis1.6-Gemma2-9B~\hfilll~\cite{Ovis} & \checkmarknew & 84.4 & 830	& \textbf{65.0} & \textbf{68.7} & \textbf{69.7} \\

 Llama-3.2-11B-Vision-Instruct~\hfilll~\cite{Llama-3.2} & \checkmarknew & 77.3 & 753 & 57.6 & 57.8 & 59.4 \\

 Pixtral-12B~\hfilll~\cite{Pixtral-12B}	 & \checkmarknew & 79.0 & 685 & 58.5 & 61.0 & 62.1 \\

 OmChat-v2.0-13B~\hfilll~\cite{OmChat} & \checkmarknew & 77.8 & 728 & 52.6 & 62.0 & 64.4 \\
 
 bailingMM-mini-17B~\hfilll~\cite{bailingmm} & \checkmarknew & 83.5 & 835 & 59.2 & 67.0 & 67.7 \\
 
 CogVLM2-19B-Chat~\hfilll~\cite{CogVLM2}		& \checkmarknew & 73.4 & 757 & 57.8 & 56.3 & 59.0 \\

 VITA-1.5-7B~\hfilll~\cite{VITA-1.5}          & \checkmarknew & 79.2 & 741 & 52.7 & 63.3 & 64.5 \\
 
 \midrule
 \multicolumn{7}{c}{Fully open models} \\

 VILA1.5-13B~\hfilll~\cite{VILA} & \crossmarknew & 69.9 & 460 & 45.0 & 49.6 & 51.5 \\

 \OurMethodBF\textbf{-16K} & \crossmarknew & 81.5 & 755 & 53.9 & \textbf{67.4} & \textbf{69.9} \\

 \OurMethodBF\textbf{-128K} & \crossmarknew & 81.1 & 738 & 53.8 & 66.9 & 69.4 \\
 
 \OurMethodBF\textbf{-1M }& \crossmarknew & 78.0 & 702 & 57.4 & 61.7 & 64.3 \\

 \bottomrule
 \end{tabular}
 \end{adjustbox}
 \end{center}
 \label{table_open_compass}
\end{table*}

\begin{table*}[!htbp]
 \caption{
 Comparison with the state-of-the-art models under \textbf{20B} parameters on Video-MME (w/o subs).
 `Internal Data' denotes whether the model is trained with in-house data, which is not publicly available.
 Most results are obtained from the leaderboard of OpenCompass.
 }
 \begin{center}
  \begin{adjustbox}{max width=0.99\textwidth}
   \begin{tabular}{l|c|c|cccc}
    \toprule

    Name
    & Internal  Data & Frames
    & Overall & Short & Medium & Long \\

    \midrule
    \multicolumn{7}{c}{Open weight models \& Partially open models} \\
    
    ARIA~\hfilll~\cite{ARIA}                        & \checkmarknew & 64 & 66.0 & \textbf{77.1} & 64.9 & 56.0 \\

    mPLUG-Owl3~\hfilll~\cite{mPLUG-Owl3}            & \checkmarknew & 16 & 54.0 & 63.3 & 51.8 & 46.8 \\

    PLLaVA-34B~\hfilll~\cite{PLLaVA}                & \checkmarknew & 16 & 53.4 & 62.0 & 52.9 & 45.4 \\
    
    InternVL2-8B~\hfilll~\cite{InternVL2}           & \checkmarknew & 16 & 53.7 & 65.9 & 49.8 & 45.3 \\
    InternVL2.5-8B~\hfilll~\cite{InternVL2.5}       & \checkmarknew & 16 & 64.2 & -- & -- & -- \\

    Qwen2-VL-7B~\hfilll~\cite{Qwen2-VL}             & \checkmarknew & 64 & 59.7 & 71.2 & 57.8 & 50.0 \\

    MiniCPM-V-2.6-7B~\hfilll~\cite{MiniCPM-V}       & \checkmarknew & 64 & 59.7 & 70.4 & 58.1 & 50.4 \\

    Idefics3-8B-Llama3~\hfilll~\cite{Idefics3}      & \checkmarknew & 16 & 54.0 & 56.1 & 45.1 & 43.0 \\

    NVILA-8B~\hfilll~\cite{NVILA}                   & \checkmarknew & 256 & 64.2 & 75.7 & 62.2 & 54.8 \\

    \midrule
    \multicolumn{7}{c}{Fully open models} \\

    LLaVA-Video-7B-Qwen2~\hfilll~\cite{LLaVA-Video} & \crossmarknew & 64 & 63.7 & 76.7 & 62.2 & 52.2 \\

    LongVILA-7B~\hfilll~\cite{LongVILA}             & \crossmarknew & 256 & 60.1 & 69.0 & 58.3 & 53.0 \\

    \midrule
    \multirow{2}{*}{\OurMethodBF\textbf{-16K}}         & \crossmarknew & 64 & 62.8 & 74.7 & 59.1 & 54.7 \\
                                            & \crossmarknew & 128 & 64.5 & 74.3 & 63.2 & 56.0 \\

    \midrule
    \multirow{5}{*}{\OurMethodBF\textbf{-128K}}     & \crossmarknew & 64 & 65.6 & 75.0 & 65.7 & 56.0 \\
                                                    & \crossmarknew & 128 & 66.2 & 74.8 & \textbf{66.7} & 57.2 \\
                                                    & \crossmarknew & 256 & \textbf{66.4} & 74.7 & 65.9 & \textbf{58.8} \\
                                                    & \crossmarknew & 512 & 65.7 & 74.7 & 64.6 & 58.0 \\

    \midrule
    \multirow{7}{*}{\OurMethodBF\textbf{-1M}}       & \crossmarknew & 64 & 59.6 & 69.2 & 57.4 & 52.0 \\
                                                    & \crossmarknew & 128 & 60.0 & 68.2 & 59.1 & 52.6 \\
                                                    & \crossmarknew & 256 & 60.7 & 68.6 & 59.7 & 53.8 \\
                                                    & \crossmarknew & 512 & 59.0 & 68.1 & 57.1 & 51.7 \\
                                                    & \crossmarknew &1024 & 57.9 & 68.4 & 57.3 & 48.0 \\
                                                    & \crossmarknew & 2048 & 56.0 & 68.2 & 57.0 & 42.7 \\
                                                    & \crossmarknew & 4096 & 55.8 & 68.2 & 56.7 & 42.6 \\

    \bottomrule
   \end{tabular}
  \end{adjustbox}
 \end{center}
 \label{table_video_mme}
\end{table*}

\begin{table*}[!htbp]
 \caption{
 Comparison with the state-of-the-art models under \textbf{20B} parameters on video benchmark.
 `Internal Data' denotes whether the model is trained with in-house data, which is not publicly available. 
 }
 \begin{center}
  \begin{adjustbox}{max width=0.99\textwidth}
   \begin{tabular}{lc|cccccccc|cc}
    \toprule

    Name
    & Internal  Data & Frames
    & LongVideoBench & MVBench \\

    \midrule

    mPLUG-Owl3-7B~\hfilll~\cite{mPLUG-Owl3}          & \crossmarknew & 128 & 52.1 & 54.5 \\
    
    LLaVA-Video-7B-Qwen2~\hfilll~\cite{LLaVA-Video} & \crossmarknew & 64 & 58.2 & 62.1 \\

    InternVL2-8B~\hfilll~\cite{InternVL2}           & \checkmarknew & 16 & 54.6 & 56.5 \\
    InternVL2.5-8B~\hfilll~\cite{InternVL2.5}  	    & \checkmarknew & 16 & 60.0 & 64.5 \\

    Qwen2-VL-7B~\hfilll~\cite{Qwen2-VL}             & \checkmarknew & 64 & 55.6 & 52.0 \\

    MiniCPM-V-2.6~\hfilll~\cite{MiniCPM-V} 	        & \checkmarknew & 64 & 54.9 & 44.7 \\

    Idefics3-8B-Llama3~\hfilll~\cite{Idefics3}	    & \checkmarknew & 16 & -- & 46.1 \\

    NVILA-8B~\hfilll~\cite{NVILA}	                & \checkmarknew & 256 & 57.7 & -- \\

    LongVILA-7B~\hfilll~\cite{LongVILA}	            & \checkmarknew & 256 & 57.1 & \textbf{67.1} \\
    
    \midrule
    \multirow{2}{*}{\OurMethodBF\textbf{-16K}}      & \crossmarknew & 64 & 59.4 & 56.6 \\
                                                    & \crossmarknew & 128 & 59.8 & 57.4 \\

    \midrule
    \multirow{4}{*}{\OurMethodBF\textbf{-128K}}     & \crossmarknew & 64 &  59.2 &	57.4 \\
                                                    & \crossmarknew & 128 & 60.7 &	57.5 \\
                                                    & \crossmarknew & 256 & \textbf{60.9} & 55.4 \\
                                                    & \crossmarknew & 512 & 59.8 & 57.3 \\

    \midrule
    \multirow{6}{*}{\OurMethodBF\textbf{-1M}}       & \crossmarknew & 64 & 53.9 & 44.7 \\
                                                    & \crossmarknew & 128 & 55.2 & 44.8 \\
                                                    & \crossmarknew & 256 & 54.0 & 44.7 \\
                                                    & \crossmarknew & 512 & 53.1 & 44.7 \\
                                                    & \crossmarknew & 1,024 & 51.8 & 44.7 \\
                                                    & \crossmarknew & 2,048 & 51.8 & 44.5 \\

    \bottomrule
   \end{tabular}
  \end{adjustbox}
 \end{center}
 \label{table_open_compass_video}
\end{table*}

\subsection{Hyper-parameters and Infrastructures}

%
%
As shown in Tab.~\ref{table_train}, we list the detailed hyperparameters in \OurMethod.

\textbf{Training.}
We configure different distributed training strategies for each key module in \OurMethod.

\begin{itemize}[leftmargin=2.0em]
    \item
    \textbf{Large Language Model.}
    We employ data, pipeline, tensor, sequence, and context parallelism.
    We enable distribution attention for context parallelism to train long sequences in Stages 3 and 4.

    \item
    \textbf{Vision Encoder.}
    We apply data, tensor, and sequence parallelism to the vision module, which is in the first LLM pipeline parallelism stage.
    We do not use context parallelism for the vision encoder.

    \item
    \textbf{Vision-Language Projector.}
    The multi-modal projector follows the configuration of the vision encoder during the distributed training.
\end{itemize}

\textbf{Inference.}
We implement two new designs to scale up the number of tokens for model inference.

\begin{itemize}[leftmargin=2.0em]
    \item
    \textbf{Context-Parallelism Distributed Inference.}
    We implement tensor parallelism with context parallelism for model inference, thus supporting distribution attention for infinite-length input tokens.
    Similar to the training mode, the length of inference tokens is fixed during the decoding phase.
    Specifically, we concatenate the input tokens with padding tokens of the maximum output length.
    The system needs to extract the new predicted next token in the fixed-length output tokens and accordingly terminate the generation process during each forward.
    \item
    \textbf{Logits-Masked Language Modeling Head.}
    We observe that the output logits from the language modeling head induce excessive memory footprints.
    For example, given $1$M tokens with $10^5$ vocabularies, the output logit matrix has a shape of $10^6 \times 10^5$ and requires $400$ GB of memory for the float32 data type.
    To address the memory issue, we mask out all hidden features and only feed the one hidden feature that predicts the next tokens to the language modeling head.
    With the above design, the memory consumption of the output logit matrix is $0.0004$ GB with $10^6 \times$ reduction.
    Note that this design can also apply to model training with long-context inputs and the language modeling head only needs to predict the short-context outputs to reduce memory consumption.
\end{itemize}
We test the maximal sequence length of a fixed number of devices before they raise the out-of-memory error.
Tab.~\ref{table_max_length} summarizes the result.
Note that activation checkpointing is disabled in LongVILA~\cite{LongVILA}, while our model has a much greater number of parameters.
Tab.~\ref{table_lm_head} further shows the effectiveness of logits-masked language modeling head~(logits-masked LM head).
All methods are implemented with Flash Attention and context-parallelism distributed inference on GPU with $96$G memory and about $150$ TFLOPS for bfloat16.
Compared to the original LM head, the logits-masked LM head extends the max sequence length by $417\%$, and reduces time cost by $47.3\%$.
We also implement a chunked language modeling head~(chunked LM head), which processes tokens with a chunk length of $32,768$.
Compared to the chunked LM head, the logits-masked LM head achieves $11.3$ and $11.1$ speedup under the $1$M  and $1.6$M input lengths, respectively.
We employ Ring Attention~\cite{RingAttention} to distribute long sequences across multiple devices.
We do not use parameter-efficient fine-tuning methods or quantization strategies for both training and inference.
Additionally, the temperature is set to $0$ to guarantee consistent performance evaluation.
%
%

\section{Experiments}

\subsection{Experiment Settings.}

We evaluate \OurMethod’s performance on image and video understanding with different sequence lengths, respectively.
We perform a comprehensive evaluation on the OpenCompass benchmark, which covers visual question answering, multimodal conversation, knowledge and reasoning, OCR, and hallucination.
OpenCompass is a comprehensive collection with $8$ popular multimodal benchmarks, including MMBench~\cite{MMBench}, MMStar~\cite{MMStar}, MMMU~\cite{MMMU}, MathVista~\cite{MathVista}, HallusionBench~\cite{HallusionBench}, AI2D~\cite{AI2D}, OCRBench~\cite{OCRBench}, and MMVet~\cite{MMVet}.
We report the average scores for all collections.
We also calculate the average scores of the objective benchmark only.
We adopt Video-MME~\cite{Video-MME} as the video understanding evaluation indicator.
Video-MME is an ideal benchmark for evaluating LMM’s ability to handle long videos in real-world scenarios, given its average video duration of 1017 seconds and the inclusion of short, medium, and long subsets.
We use greedy search to ensure reproducible experimental results.
We compare \OurMethod to a set of cutting-edge models, which we group into three families:
\begin{itemize}[leftmargin=2.0em]
    \item
    \textbf{Open weight models.}
    Models are released with only their final checkpoint; little or no information about their training data and recipe is known.
    
    \item
    \textbf{Partially open models.}
    Models are released with weights, and most of the data or details necessary to reproduce them are known.

    \item
    \textbf{Fully open models.}
    The models are fully released with weights, training data, code, and evaluation, and thus can be fully reproduced.
    
\end{itemize}

\subsection{Image Evaluation}

Tab.~\ref{table_open_compass} presents the main results on the OpenCompass leaderboard, which evaluates our model on various image benchmarks to investigate image performance.
As shown in Tab.~\ref{table_open_compass}, \OurMethod-16K demonstrates superior performance among open-source models.
In most benchmarks, \OurMethod-16K exceeds Qwen2-VL-7B and InternVL2-8B.
This highlights \OurMethod-16K’s impressive capabilities in handling multi-image tasks.
\OurMethod-16K also achieves new state-of-the-art performance on MMMU and HallusionBench and outperforms strong open-source models by a notable margin.
However, the results in \OurMethod-1M fall short of \OurMethod-16K and \OurMethod-128K, since we pack samples without isolating them via attention masks during 1M training, and the potential confusion may arise from different sources of training data.
In summary, \OurMethod achieves very competitive performance compared to other models with in-house data under the 20B parameters.
This demonstrates that pure open-source data can also build robust models with strong performance.

\subsection{Video Evaluation}

\OurMethod models also show strong video understanding capabilities.
Tab.~\ref{table_video_mme} compares different models with various frame numbers on Video-MME.
In particular, the effectiveness of \OurMethod is further underscored by its performance on Video-MME.
Specifically, \OurMethod-128K with $256$ frames exceeds all other models under the 20B parameters.
It shows exceptional results, especially in tasks involving medium- to long-length videos, outperforming other fully open video models such as LLaVA-Video-7B-Qwen~\cite{LLaVA-Video} and LongVILA-7B~\cite{LongVILA}.
\OurMethod-1M supports a maximum number of $4,096$ frames as input and achieves competitive performance.
Note that \OurMethod is fully compatible with slow-fast~\cite{SF-LLaVA} and progressive pooling~\cite{progressive_pooling} strategies, which are train-free video representations to extend the visual context window.
Meanwhile, we do not adjust the scale factor of rotary position embedding during the pre-training and fine-tuning stages; thus, employing existing interpretation methods~\cite{PositionInterpolation,YaRN} can further achieve extrapolation during the inference phase.
To further demonstrate the exceptional long- and short-context understanding capability of our method, we perform benchmark experiments on LongVideoBench~\cite{LongVideoBench} and MVBench~\cite{MVBench} for long- and short-context video understanding tasks.
The results are shown in Tab.~\ref{table_open_compass_video}, highlighting our model's efficacy across varying temporal lengths.
The \OurMethod-128K model is remarkably capable of understanding long-form video content.
Specifically, our \OurMethod-128K model surpasses all existing $7$B to $20$B model series on LongVideoBench.
Furthermore, the \OurMethod-1M model displays strong performance in video understanding of $64$ to $4,096$ frames.
Remarkably, despite achieving these impressive results, \OurMethod is only learned from open-source data compared to other models.
These results reflect a significant advancement in the research community's efforts to close the performance gap with commercial models.

\section{Conclusion}
\label{sec:conclusion}

\textbf{Contributions.}
This work introduces the \OurMethod series models as a primary exploration into powerful long-context vision-language models.
Thanks to open-source data and training infrastructures, \OurMethod has excellent results compared to the cutting-edge models under $20$B parameters with in-house data and achieves new state-of-the-art performance for both image and video understanding on several benchmarks.

\textbf{Limitations.}
Despite the promising performance, there are several limitations with the current models \OurMethod.
(1) Data Filtering.
\OurMethod is trained on massive open-source data without filtering.
Therefore, data selection still leaves plenty of room for performance improvement.
(2) Long-Content Performance.
As the \OurMethod-128K outperforms \OurMethod-1M for both image and video understanding, the training pipeline and long-content data still need to be improved.

\textbf{Future Works.}
Considering the current limitations and the promising future of vision-language models, we also anticipate increasing efforts in expanding \OurMethod capabilities to encompass other modalities, such as 3D point-cloud and audio, \etc.
We believe that simultaneous advancements in the training pipeline and multi-modal capacity will soon lead to long-content models that provide a satisfying user experience.

{
    \small
    \bibliographystyle{IEEEtran}
    \bibliography{library_format}
}


\end{document}